\documentclass[letterpaper, 10 pt, conference]{ieeeconf}  

\let\labelindent\relax

\IEEEoverridecommandlockouts                              

\usepackage{amsmath, amsthm, amssymb}
\usepackage{enumerate,enumitem}
\usepackage{graphicx}
\usepackage{algorithm, algorithmic}
\usepackage{cite}
\usepackage{balance}

\newtheorem{ass}{\textbf{Assumption}}
\newtheorem{pro}{\textbf{Proposition}}

\newtheorem{dnt}{\textbf{Definition}}

\newtheorem{prb}{\textbf{Problem}}

\title{\LARGE \bf
Motion Planning and Control with Unknown Nonlinear Dynamics through Predicted Reachability
}

\author{Zhiquan Zhang$^{\dagger}$, Gokul Puthumanaillam$^{\dagger}$, Manav Vora$^{\dagger}$ and Melkior Ornik$^{\dagger}$
\thanks{This research was supported in part by NASA under grant number 80NSSC22M0070 and by the Air Force Office of Scientific Research under grant number FA9550-23-1-0131.}
\thanks{$^{\dagger}$Zhiquan Zhang, Gokul Puthumanaillam, Manav Vora and Melkior Ornik are with the University of Illinois Urbana-Champaign, IL, 61801 USA. Emails:
        {\tt\small \{zz121, gokulp2, mkvora2, mornik\}@illinois.edu}}%
}

\begin{document}

\maketitle
\thispagestyle{empty}
\pagestyle{empty}

\begin{abstract}

Autonomous motion planning under unknown nonlinear dynamics presents significant challenges. An agent needs to continuously explore the system dynamics to acquire its properties, such as reachability, in order to guide system navigation adaptively. In this paper, we propose a hybrid planning-control framework designed to compute a feasible trajectory toward a target. Our approach involves partitioning the state space and approximating the system by a piecewise affine (PWA) system with constrained control inputs. By abstracting the PWA system into a directed weighted graph, we incrementally update the existence of its edges via affine system identification and reach control theory, introducing a predictive reachability condition by exploiting prior information of the unknown dynamics. Heuristic weights are assigned to edges based on whether their existence is certain or remains indeterminate. Consequently, we propose a framework that adaptively collects and analyzes data during mission execution, continually updates the predictive graph, and synthesizes a controller online based on the graph search outcomes. We demonstrate the efficacy of our approach through simulation scenarios involving a mobile robot operating in unknown terrains, with its unknown dynamics abstracted as a single integrator model.

\end{abstract}

\section{INTRODUCTION}
Robotic motion planning and control in unknown environments pose significant challenges across domains, including mobile robot and drone navigation \cite{shim2005autonomous, zhou2022swarm, mustafa2019towards, zhang2024efp}, planetary exploration \cite{rockenbauer2024traversing}, indoor robot exploration \cite{wang2019efficient}, etc. Robots need to operate without complete or precise knowledge of the surroundings and might need to contend with limited sensing range, risky and unstructured dynamics, and finite computational resources \cite{dixit2024step, ahmadi2020risk, ono2015chance}.

Several studies have addressed the challenge of exploring unknown environments while fulfilling high-level task specifications. For instance, \cite{ayala2013temporal} introduced a motion planning framework that employs frontier-based exploration coupled with an incremental search algorithm to determine a feasible path that fulfills a predefined task, separating the phases of exploring and exploiting. The work of \cite{nawaz2020explorative} integrated these phases by performing search on a probabilistic graph that leverages prior knowledge of the environment, thereby blurring the traditional distinctions between exploration and exploitation. However, these approaches represent environmental uncertainty as unknown state labels, neglecting the uncertainties associated with low-level environmental and robotic dynamics, and control synthesis under uncertainties in dynamics.

This paper focuses on scenarios in which a robot seeks to reach a target location while operating under unknown dynamics (dictated by both its mechanics and uncertain environments). The challenge arises partly from the difficulty in conducting reachability analysis under unknown dynamics with constrained control inputs \cite{shafa2022reachability}, as well as from devising an ``optimal'' strategy that effectively balances identifying the dynamics (exploration) and synthesizing controllers (exploitation) during state space search.

In the context of control under unknown dynamics, the control community frequently adopts data-driven approaches for online system identification and control synthesis. For example, \cite{zhang2011data} employed a reinforcement learning framework to develop an approximate optimal control strategy that drives the system from any initial state to zero, with rigorous proof for Lyapunov stability. Additionally, adaptive control methods \cite{landau2011adaptive, bar1988simple} are utilized to design control strategies while concurrently learning the system dynamics online. However, these approaches primarily focus on parameter learning within structured system models and presuppose the reachability of the target state. Furthermore, many existing approaches are concerned with achieving the target state over an infinite time horizon, without addressing the trajectory characteristics, thus limiting their applicability to complex and potentially interactive robotic task specifications.

In our work, we tackle the motion planning problem under both reach specifications and unknown low-level dynamics by proposing a hybrid planning-control framework. We partition the state space into polytopic regions and approximate the dynamics within each polytope as an affine system, subsequently abstracting the system's transitions between polytopes by a directed weighted graph. The edges of the graph, which are initially unknown, are updated incrementally by evaluating the reachability between polytopes (i.e. vertices of the graph), enabling adaptive search for feasible paths as the agent concurrently explores and exploits. The dynamics within a polytope are identified once the agent has entered. Notably, our approach determines reachability not only for regions whose dynamics have already been identified, but also for those with unknown dynamics by introducing a predictive condition based on available prior system information. The core idea of our work is similar to that in \cite{nawaz2020explorative}. However, our approach characterizes environmental uncertainty in terms of uncertain robotic and environmental dynamics rather than as unknown target locations. Our work is also closely related to that of \cite{meng2024online}, which also addresses reachability while synthesizing controllers in unknown dynamics, however their approach requires extensive data and involves significant computational complexity.

The rest of this paper is organized as follows. Section \ref{sec:pre} introduces necessary mathematical preliminaries and notations. Section \ref{sec: PF} presents formal formulation of the problem. Section \ref{sec:approach} proposes our hybrid motion planning framework which integrates both path planning and control synthesis. Section \ref{sec:experiment} demonstrates the framework through experiments involving a model of a ground mobile robot operating on unknown and challenging terrain.

\section{PRELIMINARIES AND NOTATIONS} \label{sec:pre}
We start by introducing some preliminaries and notations related to transition systems, directed weighted graphs, polytopes, mathematical notations and terminology. 

\subsection{Transition Systems and Directed Weighted Graphs}
A deterministic transition system \cite{baier2008principles} is defined as a six-tuple $\mathcal{TS} = (S, Act, \rightarrow, I, AP, L)$, where $S$ is a non-empty set whose elements $s\in S$ are \textit{states}. A non-empty set $Act$ whose elements $a$ are referred to as \textit{actions} or \textit{events}. $\rightarrow \subseteq S \times Act \times S$ is a ternary relation. The expression $(s, a, s') \in \rightarrow$ indicates the system transition from state $s$ to state $s'$ by performing action $a$. The set of all possible actions is denoted by $Act$. The set $AP$ is a set of \textit{atomic propositions} used to describe properties of the states. \textit{Labeling function} $L: S \rightarrow 2^{AP}$ is a function that assigns to each state $s \in S$ a set of atomic propositions $L(s)\subseteq AP$ that hold true in that state.

A \textit{path} (or \textit{run}) starting from an initial state $s_0 \in I$ is a sequence of states and actions $\pi = s_0 \stackrel{a_0}{\longrightarrow}s_1\stackrel{a_1}{\longrightarrow}s_2\stackrel{a_2}{\longrightarrow}\ldots$, where for all $i \ge 0,\ (s_i, a_i, s_{i+1}) \in \rightarrow$. A state $s' \in S$ is said to be reachable if there exists an initial state $s_0 \in I$ and a path starting from $s_0$ such that $s'$ appears as one of the states in the path.

In this paper, a transition system is associated with a \textit{directed weighted graph} $\mathcal{G} = (V, E, w, AP, L)$. The set of \textit{vertices} $V$ of the graph is the set of states $S$ of the transition system. We define the set of \textit{edges} $E$ as those pairs of states $s, s'$ for which there exists an action $a \in Act$ such that $(s, a, s') \in \rightarrow$. Suppose that the transitions are associated with a weight function $c:E \rightarrow \mathbb{R}$, which assigns a weight to each edge. We define a weight function $w: E \rightarrow \mathbb{R}$ on the graph as $w(s, s') = c(e),\ e\in E, s, s'\in V$. For a path $\pi$ in the transition system, the corresponding weight of the path in the graph becomes $w(\pi) = \sum_iw(s_i, s_{i+1}) = \sum_i c(e)$. The labeling function $L$ and the set of atomic propositions $AP$ are defined in the same manner as the corresponding transition system.

\subsection{Polytopes}
Polytopes can be described by the following two equivalent definitions \cite{ziegler2012lectures}:
\begin{dnt}
    A set $\mathcal{P} \subseteq \mathbb{R}^n$ is called a polytope if there exists a finite set of points $\{v_1, v_2, \cdots, v_m\} \subset \mathbb{R}^n$ ($m \ge n+1$) such that
    \begin{equation}\small
        \mathcal{P} = {\rm conv}\{v_1, v_2, \cdots, v_m\} = \left\{\sum_{i=1}^m \lambda_i v_i:\lambda \ge 0, \sum_{i=1}^m \lambda_i = 1 \right\}.
    \end{equation}
\end{dnt}
That is, a polytope is the convex hull of finitely many points.
\begin{dnt}
    A set $\mathcal{P} \subseteq \mathbb{R}^n$ is a polytope if and only if there exists a matrix $A\in \mathbb{R}^{k \times n}$ and a vector $b \in \mathbb{R}^k$ ($k \ge n+1$) such that
    \begin{equation}
        \mathcal{P} = \{x\in \mathbb{R}^n:Ax \leq b\},
    \end{equation}
    and $\mathcal{P}$ is bounded.
\end{dnt}

We only consider convex polytopes in this paper. A hyperplane $H \subseteq\mathbb{R}^n$ is said to be a \textit{supporting hyperplane} of $\mathcal{P}$ if $\mathcal{P}$ is contained entirely in one of the closed half-spaces determined by $H$ and $\mathcal{P} \cap H\neq \varnothing$. A subset $F \subseteq \mathcal{P}$ is called a \textit{face} of $P$ if there exists a supporting hyperplane $H$ of $\mathcal{P}$ such that $F = \mathcal{P} \cap H$. A \textit{facet} is a face of $\mathcal{P}$ of dimension $n-1$. A \textit{vertex} is a face of dimension $0$. Equivalently, a point $v\in P$ is a vertex of $P$ if whenever $v = \lambda x + (1-\lambda)y,\ x, y\in P,\ \lambda\in [0, 1]$, $x = y = v$ is indicated.

\subsection{Notations and Terminology}
For any function $f(x)$, $\nabla_x f(x)$ denotes $\partial f(x)/\partial x$. For any vector $v$, $\|v\|$ denotes the Euclidean norm and for any matrix $M$, $\|M\|$ denotes $\sup_{x \neq 0}(\|Mx\|/\|x\|)$, where $x$ is any vector. 

For clarity in the subsequent sections, we define the term ``facet reachability'' from a \textit{reach control} \cite{habets2004control, broucke2014reach} perspective. A facet $F$ of $\mathcal{P}$ is \textit{reachable} under affine dynamics if, for any state $x_0$ in $\mathcal{P}$, there exist a time $T_0>0$ and an affine feedback control law such that:
\begin{enumerate}
    \item the state trajectory remains within $\mathcal{P}$ for all $t \in [0, T_0)$, i.e., $x(t)\in \mathcal{P}$; 
    \item the state reaches the facet $F$ of $\mathcal{P}$ at time $T_0$, i.e., $x(T_0) \in F$; 
    \item the velocity direction coincides with the outward normal vector $n$ of $F$, i.e., $n^\top\dot x(T_0) > 0$.
\end{enumerate}
Given that reach control with continuous feedback remains an open question \cite{ornik2015topological}, we focus on constructing affine feedback control for simplicity in this paper.

\section{PROBLEM FORMULATION} \label{sec: PF}
We consider an agent operating in an unknown continuous-time nonlinear control-affine system
\begin{equation}\label{dynamical system}
    \dot x = f(x) + g(x)u,
\end{equation}
where $x \in \mathbb{R}^n$ denotes the system state, $f(x)\in\mathbb{R}^n \rightarrow \mathbb{R}^n$, $g(x) \in \mathbb{R}^n \rightarrow \mathbb{R}^{n\times m}$ denote the drift dynamics and the control dynamics respectively. $u \in \mathbb{R}^m$ is the control input. We make the following assumptions to formally define the problem.
\begin{ass}\label{polytope ass}
    \textbf{(State-space and control constraints)} The state space of the is constrained by a polytope $P_s \subseteq \mathbb{R}^n$, and the control input is constrained by a polytope $P_u \subseteq \mathbb{R}^m$.
\end{ass}
\begin{ass}\label{Lip ass}
    \textbf{(Lipschitz continuity)} Functions $\nabla_x f(x)$ and $g(x)$ (\ref{dynamical system}) are Lipschitz continuous on $P_s$. Namely, there exist constants $\mathcal{L}_{df}$ and $\mathcal{L}_g$ such that $\|\nabla_xf(x_1) - \nabla_xf(x_2)\| \leq \mathcal{L}_{df}\|x_1 - x_2\|$ and $\|g(x_1) - g(x_2)\| \leq \mathcal{L}_g\|x_1 - x_2\|$ for any $x_1, x_2\in P_s$.
\end{ass}
Assumption \ref{Lip ass} holds for any functions $f(x)$ and $g(x)$ in $\mathcal{C}^2$, given that $P_s$ is compact \cite{bartle2000introduction}. This assumption assures that, for a sufficiently smooth control input $u$, system \eqref{dynamical system} results in existence and uniqueness of trajectories.

In this paper, we consider the following problem:
\begin{prb}
    Consider an agent operating in a nonlinear system described by (\ref{dynamical system}), where functions $f(x)$ and $g(x)$ are unknown, but their Lipschitz constants $L_{df}$ and $L_g$ are given. Given an initial state $x(0) = x_0, x_0\in P_s$ and $x^*\in P_s$, the objective is to determine a control signal $u$, such that the agent reaches $x^*$ within finite time.
\end{prb}

This problem is inherently challenging due to the unknown nature of the dynamics, which introduce significant difficulties in synthesizing feasible control inputs, relying on performing rigorous reachability analysis and leveraging system identification (exploration) and motion planning and control synthesis (exploitation). 

To address these challenges, our approach involves two steps: (1) approximating the dynamics by piece-wise affine system, and (2) formulating the version of Problem 1 tailored to these approximated dynamics described in Problem 2. We firstly partition the state space $P_s$ into a finite collection of disjoint polytopes $\{P_l\}_{\ l \in L}$, where $L$ is a finite index set. Within each polytope, the system dynamics are approximated by an affine model, yielding an overall continuous-time piecewise-affine system. The facets of these polytopes represent the boundaries at which the system switches between different affine regimes. The linearization of \eqref{dynamical system} at $(x_e, u_e = 0)$, where $x_e$ is the center of $P_l$, is given by
\begin{equation}\label{piecewise affine}
    \begin{aligned}
        f(x)+g(x)u \approx& f(x_e) + g(x_e)u_e + \nabla_x f(x_e)(x-x_e) \\
        &+ \nabla_x g(x_e)(x-x_e)u_e + g(x_e)(u-u_e)\\
        =&\nabla_x f(x_e) x + g(x_e)u + f(x_e) - \nabla_x f(x_e)x_e\\
        =&\bar A_l x + \bar B_l u + \bar c_l,
    \end{aligned}
\end{equation}
where $\bar A_l = \nabla_x f(x_e) \in \mathbb{R}^{n\times n}$, $\bar B_l = g(x_e) \in \mathbb{R}^{n\times m}$, and $\bar c_l = f(x_e) - \nabla_x f(x_e)x_e \in \mathbb{R}^n$ represent the approximated affine system dynamics within the polytope $P_l$ \cite{Khalil:1173048, ornik2017automated}. Obviously, if the partition of $P_s$ into polytopes $P_l$ is sufficiently fine, the maximal difference between $f(x)+g(x)u$ and $\bar A_l x + \bar B_l x + \bar c_l$ as defined in \eqref{piecewise affine} can be made arbitrarily small.

 We associate the partitioned state space with a directed weighted graph $\mathcal{G}_s = (V_s, E_s, w_s)$. The vertex set is defined as the index set $L$ of the polytopes. A directed edge $(l, l')\in E_s$ is included if the common facet  $F_{l,l'}$ between polytope $P_l$ and $P_{l'}$ is reachable for any initial state $x_{l0}\in P_{l}$. Let $P_i$ and $P_*$ denote the polytopes containing the initial state $x_0$ and the target state $x^*$ respectively. Importantly, the unknown nature of the dynamics implies uncertainty of reachability, which means the existence of edges $E_s$ is not known to the agent at the beginning of its mission. However, since the partitioning of the state space is predefined, the vertex set $V_s$ is known.

We cast the original problem as a joint path-planning and control synthesis task over a partitioned state space represented by the directed weighted graph $\mathcal{G}_s$. This leads to the following problem formulation.
\begin{prb}
    Consider an agent operating on an unknown nonlinear system which is piece-wise approximated by \eqref{piecewise affine}, where the state space is partitioned into disjoint polytopes. Synthesize and affine feedback controller to drive the agent from the initial polytope $P_i$ to the target polytope $P_*$.
\end{prb}

This formulation provides a tractable framework for addressing the challenges associated with driving the system from $x_0$ to $x^*$ via piecewise-affine approximations.

\section{PROPOSED APPROACH} \label{sec:approach}
Our proposed approach adopts a hierarchical framework consisting of two complementary layers. The low-level layer focuses on dynamics identification and polytope-scale reachability analysis while the high-level layer performs search on a graph abstracted from the state space partition and dynamics linearization. In this structure, affine system identification and reachability analysis continuously update the graph's edges, and graph search subsequently determines the regions that the agent should explore.

Subsection \ref{subsec:reacha} restates the reachability condition and controller synthesis procedure as introduced in \cite{habets2004control}. It also predicts the facet reachability associated with polytopes with unknown dynamics using prior system knowledge. Subsection \ref{subsec: RGPG} provides details of the construction of the graph, with heuristic weights assigned to edges with both certain and uncertain existence, allowing the agent to balance the trade-off between exploration and exploitation. Subsection \ref{subsec: motion} presents a unified motion planning framework that guides the selection of subsequent polytopes based on the current polytope, while synthesizing the low-level controllers to execute the plan.

\subsection{Facet Reachability and Controller Synthesis}\label{subsec:reacha}
Suppose that at time $t_0$ the state of the agent lies within polytope $P_l$, i.e., $x(t_0) \in P_l$. Let $v_{l1}, \cdots, v_{lM}$ denote the vertices of $P_l$, and let $F_{l1}, \cdots, F_{lK}$ denote the facets of $P_l$, associated with normal vectors $n_{l1}, \cdots, n_{lK}$ pointing outward $P_l$. We define $V_{li}\subset\{1, \cdots, M\}$ as an index set of vertices of the facet $F_{li}$, i.e., $\{v_{lj}|j\in V_{li}\}$. We also define $W_{lj} \subset\{1, \cdots, K\}$ as an index set of facets which contain $v_{lj}$. Assume that the approximate affine system has been identified as described in Appendix B once it has entered $P_l$ and is represented by $(\bar A_l, \bar B_l, \bar c_l)$. In order to describe the agent transition from one polytope to another, without any loss of generality, we focus on the reachability of facet $F_{l1}$ (with normal vector $n_{l1}$). Suppose there exist control inputs $u_{l1}, \cdots, u_{lM} \in P_u$ and the following conditions are satisfied:
\begin{enumerate}
    \item $\forall j \in V_1$:
    \begin{enumerate}[label=(\theenumi\alph*)]
        \item $n_1^\top (\bar A_l v_j + \bar B_l u_j + \bar c_l)>0$;
        \item $\forall i \in W_j \backslash\{1\}:\ n_i^\top (\bar Av_j + \bar Bu_j +\bar c_l)\leq 0$;
    \end{enumerate}
    \item $\forall j \in \{1, \cdots, M\}\backslash V_1$:
    \begin{enumerate}[label=(\theenumi\alph*)]
        \item $\forall i \in W_j:\ n_i^\top (\bar A_l v_j + \bar B_l v_j + \bar c_l) \leq 0$;
        \item $n_1^\top (\bar A_l v_j + \bar B_l u_j + \bar c_l)>0$.
    \end{enumerate}
\end{enumerate}
Under these conditions, an affine feedback controller can be synthesized to drive the system to reach the facet $F_{l1}$ in finite time. The procedure, introduced in \cite{habets2004control, roszak2005necessary}, is as follows:

We begin by determining a set of feasible control inputs $u_{l1}, u_{l2}, \cdots, u_{lM}$ that satisfy the constraints (1a)-(2b). Then perform a triangulation of the polytope $P_l$ and select a simplex $\bar P_l \subset P_l$ whose vertices $\bar v_{l1}, \bar v_{l2}, \cdots, \bar v_{l(n+1)}$ are each associated with a control input $\bar u_{l1}, \bar u_{l2}, \cdots, \bar u_{l(n+1)}$. Subsequently, we construct an affine feedback control law of the form
\begin{equation}
    u = F_lx + g_l,
\end{equation}
where $F_l$ and $g_l$ are obtained by solving
\begin{equation}\label{eq: controller}
    \begin{bmatrix}
        F_l|g_l
    \end{bmatrix}\begin{bmatrix}
        v_{l1}&\cdots & v_{l(n+1)}\\
        1&\cdots &1
    \end{bmatrix} = \begin{bmatrix}
        \bar u_{l1}&\cdots & \bar u_{l(n+1)}
    \end{bmatrix}.
\end{equation}

Determining whether there exists a feasible set of control inputs $(u_{l1}, u_{l2}, \cdots, u_{lm})$ that satisfies the specified conditions, and computing a corresponding solution, amounts to finding a feasible point within a polytope defined by the intersection of the hyperplanes specified in constraints (1a)-(2b) and the control constraint polytope $P_u$.

In addition to assessing the reachability of the facets of $P_l$, we propose a predictive condition, under the Lipschitz assumption outlined in Assumption \ref{Lip ass}, to determine the facet reachability for agents located within another polytope $P_{l'}$, where the approximate affine dynamics are unknown. As discussed above, reachability relies on the feasibility of several affine constraints parameterized by $(\bar A_l, \bar B_l, \bar c_l)$. Thus, for polytopes whose dynamics are still unknown, we can utilize the bounded differences between its unknown dynamics and the identified model $(\bar A_l, \bar B_l, \bar c_l)$ to conservatively adjust the feasible range.

To provide the reachability conditions, for simplicity, we firstly categorize the above-mentioned conditions (1a)-(2b) into two groups: those expressed as strict inequalities and non-strict inequalities. We assume the affine dynamics associated with $P_{l'}$ are given by $(\bar A_{l'}, \bar B_{l'}, \bar c_{l'})$. Let $v_{l'1}, \cdots, v_{l'M}$ denote the vertices of $P_{l'}$, $F_{l'1}, \cdots, F_{l'K}$ denote its facets, each associated with outward-pointing normal vectors $n_{l'1}, \cdots, n_{l'K}$. Without loss of generality, we focus on inequalities related to a specific vertex $v_{lj}$ and its corresponding control input $u_{lj}$. Define the index set $I_+ = \{i|n_{l'i}^\top (\bar A_{l'} v_{l'j} + \bar B_{l'} u_{l'j} + \bar c_{l'})>0\}$ and the index set $I_- = \{i|n_{l'i}^\top (\bar A_{l'} v_{l'j} + \bar B_{l'} u_{l'j} + \bar c_{l'}) \leq 0\}$. Naturally, the index set $I^+$ and $I^-$ is determined by the choice of the exit facet $\bar F_{l'}$, which determines how $V_{l'i}$ and $W_{l'j}$ are arranged. Then, the condition set (1a)-(2b) is rewritten as:
\begin{equation}\label{eq: simple condition}
    \left\{
    \begin{aligned}
    n_{l'i}^\top (\bar A_{l'} v_{l'j} + \bar B_{l'} u_{l'j} + \bar c_{l'}) &> 0, i\in I_+\\
    n_{l'i}^\top (\bar A_{l'} v_{l'j} + \bar B_{l'} u_{l'j} + \bar c_{l'}) &\le 0, i\in I_-\\
    u_{l'j} &\in P_u.
    \end{aligned}
    \right.
\end{equation}

We then consider the predictive reachability conditions for the polytope $P_{l'}$, which is characterized by the unknown affine dynamics $(\bar A_{l'}, \bar B_{l'}, \bar c_{l'})$. Two distinct sufficient conditions determining the feasibility or infeasibility \eqref{eq: simple condition} are presented. 

Firstly, we present the sufficient conditions for the feasibility of \eqref{eq: simple condition} by perturbing the known dynamics $(\bar A_l, \bar B_l, \bar c_l)$ to the largest extend yielding the feasible range associated with $(\bar A_{l'}, \bar B_{l'}, \bar c_{l'})$ defined by \eqref{eq: simple condition} \underline{\textbf{maximally narrowed}} compared to that associated with $(\bar A_l, \bar B_l, \bar c_l)$. To capture this conservative scenario, we construct a \textit{robust inequality set} \eqref{narrow ineq} that involves both the identified dynamics $(\bar A_l, \bar B_l, \bar c_l)$ and dynamics difference bounds $\varepsilon_A$, $\varepsilon_B$, $\varepsilon_c$ given by Appendix A. This robust formulation represents the most conservative reachability condition and ensures that if a feasible solution exists under these tightened constraints, the agent can be guaranteed to reach the facets of $P_{l'}$.

For a $m$-dimensional control input $u_{lj} = [u_{lj}^1, u_{lj}^2, \cdots, u_{lj}^m]^\top$, which is the variable of the \textit{robust inequality set}, the sign of each component affects the structure of the robust inequality sets. To account for this, we define the index sets $U^+ = \{k|u_{lj}^k >0\}$ and $U^-=\{k|u_{lj}^k\leq 0\}$.
\begin{pro}
Consider $2^m$ sets of inequalities formulated by \eqref{narrow ineq}. 
\begin{equation}\label{narrow ineq}
    \scalebox{0.93}{$\left\{
    \begin{aligned}
    \bar B_{l'}^-  &u_{l'j} + n_{l'i}^\top (\bar A_{l} v_{l'j}  + \bar c_{l}) > \|n_{l'i}\|(\varepsilon_A\|v_{l'j}\| + \varepsilon_c), i\in I_+\\
     \bar B_{l'}^+ &u_{l'j} + n_{l'i}^\top (\bar A_{l} v_{l'j}  + \bar c_{l})  \leq- \|n_{l'i}\|(\varepsilon_A\|v_{l'j}\| + \varepsilon_c), i\in I_-\\
    &u_{l'j}^k >0,k\in U^+\\
    &u_{l'j}^{k'} \leq0, k \in U^-\\
    &u_{l'j} \in P_u,
    \end{aligned}
        \right.$}
\end{equation}
where $\varepsilon_A$, $\varepsilon_B$, $\varepsilon_c$ are bounds for dynamics difference between $(\bar A_l, \bar B_l, \bar c_l)$ and $(\bar A_{l'}, \bar B_{l'}, \bar c_{l'})$ are estimated as described in Appendix A. The vectors $ \bar B_{l'}^-$ and $\bar B_{l'}^+$ are defined by:
\begin{equation}
    \begin{aligned}
        \bar B^+_{l'} &= n_{l'i}^\top \bar B_l+ \left[\frac{u_{l'j}^1}{|u_{l'j}^1|}\varepsilon_B\|n_{l'i}\|\ \cdots \frac{u_{l'j}^m}{|u_{l'j}^m|}\varepsilon_B\|n_{l'i}\|\right ],\\
        \bar B^-_{l'} &= n_{l'i}^\top \bar B_l-\left[\frac{u_{l'j}^1}{|u_{l'j}^1|}\varepsilon_B\|n_{l'i}\|\ \cdots\ \frac{u_{l'j}^m}{|u_{l'j}^m|}\varepsilon_B\|n_{l'i}\|\right ].\\
    \end{aligned}
\end{equation}
If at least one of these sets is feasible, \eqref{eq: simple condition} is guaranteed to be feasible. This feasibility indicates that the exit facet $\bar F_{l'}$ within the polytope $P_{l'}$ is reachable.
\end{pro}
Since there are $m$ components in $u_{lj}$, this formulation results in $2^m$ distinct inequality sets that together cover the entire space of $u_{lj}$. If at least one of these sets is feasible, the scenario where the original feasible range of $u_{l'j}$ is narrowed, is therefore considered feasible. It is obvious to see that if $\varepsilon_A=\varepsilon_B = \varepsilon_c = 0$, indicating that $P_l$ and $P_{l'}$ are identical, \eqref{narrow ineq} reduces to \eqref{eq: simple condition}. It appears that \eqref{narrow ineq} leads to a smaller feasible range compared to using $(\bar A_l, \bar B_l, \bar c_l)$ as the dynamics of $P_{l'}$. See Appendix C for a detailed proof.

Analogously, we present the sufficient condition for infeasibility of \eqref{eq: simple condition} by perturbing the known dynamics $(\bar A_l, \bar B_l, \bar c_l)$ to the largest extend yielding the feasible range of $u_{l'j}$ correspond to $(\bar A_{l'}, \bar B_{l'}, \bar c_{l'})$ \underline{\textbf{maximally expanded}} compared to that associated with $(\bar A_l, \bar B_l, \bar c_l)$. Using the same notations as in Proposition 1, we propose the following proposition.
\begin{pro}
Consider $2^m$ distinct inequality sets formulated by \eqref{expand ineq}. 

\begin{equation}\label{expand ineq}
    \scalebox{0.93}{$\left\{
    \begin{aligned}
     \bar B_{l'}^+ &u_{l'j} + n_{l'i}^\top (\bar A_{l} v_{l'j}  + \bar c_{l})> -\|n_{l'i}\|(\varepsilon_A\|v_{l'j}\| + \varepsilon_c), i\in I_+\\
     \bar B_{l'}^- &u_{l'j} + n_{l'i}^\top (\bar A_{l} v_{l'j}  + \bar c_{l})  \leq  \|n_{l'i}\|(\varepsilon_A\|v_{l'j}\| + \varepsilon_c), i\in I_-\\
    &u_{l'j}^k >0,k\in U^+\\
    &u_{l'j}^{k'} \leq0, k \in U^-\\
    &u_{l'j} \in P_u,
    \end{aligned}
        \right.$}
\end{equation}

If all of these sets are infeasible, \eqref{eq: simple condition} is guaranteed to be infeasible. This indicates that the exit facet $\bar F_{l'}$ within $P_{l'}$ is unreachable.
\end{pro}

The proof is similar to that of Proposition 1 and, for brevity, is therefore omitted.
\subsection{Predictive Reachability-Guaranteed Graph}\label{subsec: RGPG}
The predictive reachability-guaranteed graph $G_s = (V_s, E_s, w_s)$ follows as a consequence of Proposition 1 and Proposition 2. Namely, a directed edge from $P_l$ to $P_{l'}$ exists if 1) $P_l$ and $P_l'$ have a common facet $F_{l,l'}$, and 2) the inequality set \eqref{narrow ineq} is feasible by treating the exit facet as $F_{l,l'}$. Conversely, an edge from $P_l$ to $P_{l'}$ does not exist if either 1) $P_l$ and $P_{l'}$ are not adjacent, or 2) the inequality set \eqref{expand ineq} is infeasible. 

For edges with certain existence, weights can be assigned based on the upper bound of time $T_0$ required for the agent to reach the corresponding facet. Consider a set of inputs $u_{l1}, \ldots, u_{lM}$ and the corresponding polytope vertices $v_{l1}, \ldots, v_{lM}$. Without loss of generality, assume that the agent entering polytope $P_l$ at $x_{l0}$ aims to reach the facet $n_{l1}$, the upper bound of $T_0$ is given by $T_0\leq (\beta-\alpha)/c_1$, where $\alpha = n_1^\top x_{l0}$, $\beta =  \max\{n_1^\top v_{lj}|j=1,\cdots,m\}$, and $c_1 = \min\{n_1^\top(\bar A_l v_{lj} + \bar B_l u_{lj}+\bar c_l)|j=1,\cdots,M\}$. Please refer to \cite{habets2004control} for detailed proof.

In the case where the inequality set \eqref{narrow ineq} is infeasible while the inequality set \eqref{expand ineq} is feasible, the existence of an edge remains indeterminate, and we include them in the graph, but assign heuristic weights. The assigned heuristic weights are based on their exploratory potential. Let $V_{explored}$ denote the set of polytopes for which the dynamics has been identified, and let $P_{explored}^i$ be the $i$-th polytope in $V_{explored}$. For a directed edge $e_u\in E_s$ from $P_l$ to $P_{l'}$ whose existence is uncertain, we define its weight $w_u$ as:
\begin{equation}\label{eq:unknown weight}
    w_u = \gamma \bar w_e\frac{\sum_{i=1}^{\#(V_{explored})}\frac{1}{d(P_{l'}, P_{explored}^i)}}{\#(V_{explored})},
\end{equation}
where $\#(V_{explored})$ is the cardinality of $V_{explored}$, $d(P_{l'}, P_{explored}^i)$ is the Euclidean distance between the center of $P_{l'}$ and that of $P_{explored}^i$, $\bar w_e$ is the average of weights assigned to known edges, and $\gamma>0$ is a constant to set the scale of $w_u$ with respect to that of $\bar w_e$. A smaller $\gamma$ indicates that we place greater emphasis on the impact of known edges on the graph search. This formulation serves as a metric for the average distance between $P_{l'}$ and regions where the dynamics are already known. By employing this weighting scheme, the agent is encouraged to prioritize exploration in areas that are likely to yield significant new information, potentially resulting in a shorter overall path.

\subsection{Motion Planning Framework}\label{subsec: motion}
Algorithm \ref{alg:motionplan} illustrates the motion planning framework. Initially, the robot's state space is partitioned, and mapped to a predictive reachability-guaranteed graph $G_s$ with designated initial and target nodes. Next, an iterative loop is executed that integrates high-level graph search with low-level controller synthesis. In each iteration, the affine dynamics associated with the current polytope are identified by Algorithm 2 and assumed to represent the entire polytope. Subsequently, the reachability from each polytope to its neighbors is evaluated, including those with unknown dynamics. Based on these evaluations, $G_s$ is updated by removing infeasible edges, including feasible edges, and assigning weights $w_u$ to uncertain edges $w_e$ to known edges. We then employ Dijkstra's algorithm \cite{cormen2022introduction} to determine the shortest path on $G_s$ to the target node, and consider the first following vertex on this path, denoted by $Path(1)$. We then synthesize an affine feedback controller using \eqref{eq: controller}, driving the agent to the polytope indicated by $Path(1)$. This exploration-navigation loop terminates if the agent enters the target polytope.

Naturally, the approximated affine dynamics might occasionally lead the agent to an unintended polytope. Furthermore, given that the dynamics are unknown, and we do not assume that the target is reachable even under known dynamics, it is possible for the agent to become stuck in a polytope or fail to progress toward the target.
\begin{algorithm}
\caption{Motion Planning Framework}
\begin{algorithmic}[1]\label{alg:motionplan}
\REQUIRE Lipschitz Constants $L_{df},L_g$

\STATE \textbf{Initialize:}
\[
\begin{array}{l}
 \text{State Space Partition with Polytopes}\\
 Node_{\text{initial}},\ \ Node_{\text{target}},\ \ Node_{\text{current}} \gets Node_{\text{initial}}\\
 Node_{\text{explored}} \gets \text{empty list}
\end{array}
\]
\WHILE{$Node_{\text{current}} \neq Node_{\text{target}}$}
    \STATE $[A,B,c] \gets \text{System Identification}\ (Node_{\text{current}})$
    \STATE Append $Node_{\text{current}}$ to $Node_{\text{explored}}$
    \STATE $\text{Update $G_s$}\ (A, B, c, L_{df}, L_g, Node_{\text{explored}})$
    \STATE $\text{Path} \gets \text{Graph Search ($G_s$, $Node_{\text{current}}$, $Node_{\text{target}})$}$
    \STATE $Node_{\text{current}} \gets \text{Path(1)}$
    \STATE Synthesize Controller to Reach $Node_{\text{current}}$
\ENDWHILE
\end{algorithmic}
\end{algorithm}
\begin{figure*}[t]
  \centering
  \includegraphics[width=\textwidth]{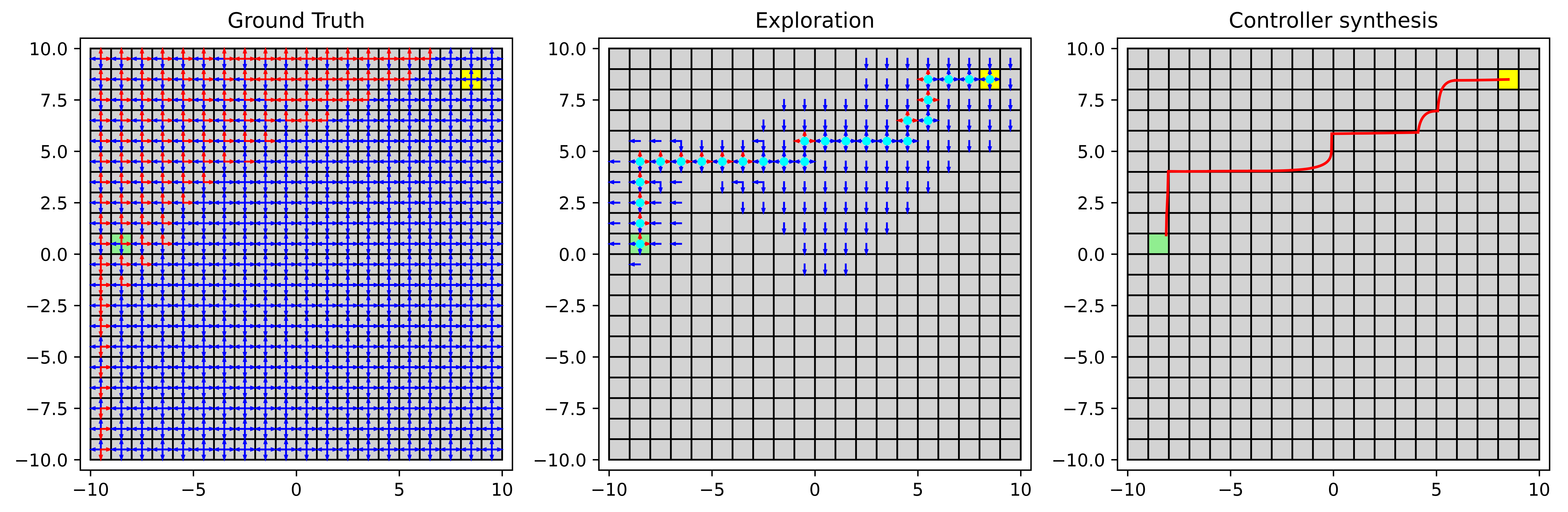} 
  \caption{The initial tile is marked yellow while the target tile is marked green. (left) The ground truth of possible transitions between polytopes. The dynamics are linearized at the center of each grid cell according to \eqref{piecewise affine}. (center) The exploration and navigation process. The sequence of the bright blue dots represents the result of graph planning. Blue arrows denote guaranteed facet reachability (i.e., the confirmed presence of edges until the mission is completed), red arrows indicate the absence of an edge and the lack of an arrow denotes uncertainty regarding edge existence. (right) The trajectory generated by the controller synthesized by \eqref{eq: controller}. }
  \label{fig:1}
\end{figure*}

\begin{figure}[t]
  \centering
  \includegraphics[width=0.47\textwidth]{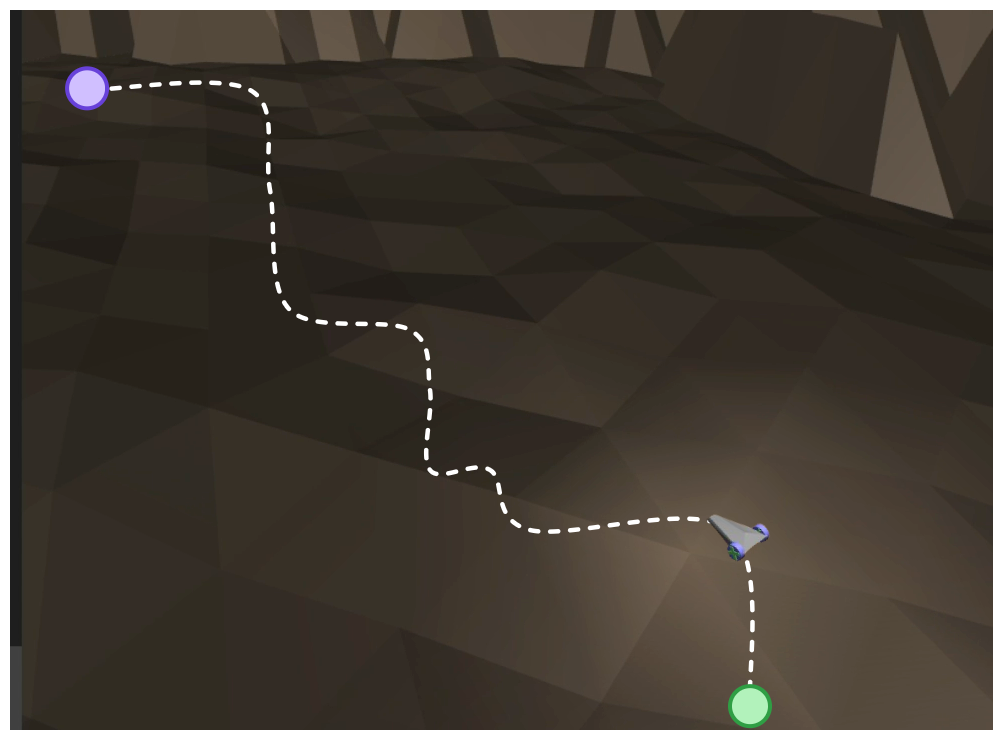} 
  \caption{The trajectory of the mobile robot. The green and purple markers denote the initial and target locations respectively.}
  \label{fig:2}
\end{figure}
\section{EXPERIMENTAL RESULTS}\label{sec:experiment}
In this section, we showcase a simulation to illustrate the efficacy of our approach, which involves a mobile robot with unknown dynamics operating in unknown terrains.

The true dynamics of the mobile robot operating in unknown terrains is approximated using a single integrator model:
\begin{equation}
\begin{aligned}
    \dot x &= f(x)+g(x)u,\\
    f(x) &= \begin{bmatrix}
        -0.5\sin(0.1x_1-0.2x_2)-4.5\\
         -0.2\sin(0.3x_1-0.1x_2)-4.5
    \end{bmatrix},\\
    g(x) &= \begin{bmatrix}
        1+0.02x_1& 0.02x_2\\
        -0.02x_1& 1-0.02x_2\\
    \end{bmatrix},
\end{aligned}
\end{equation}
which is treated as unknown, except it is known that its Lipschitz constants are $L_{df}=0.03$ and $L_{g}=0.03$. The drift dynamics $f(x)$ capture sinusoid perturbations while the control dynamics $g(x)$ represent a state-dependent control input. The two-dimensional state space is partitioned into a 20-by-20 grid world with equal-sized tiles, constrained by $-10\leq x_1\leq 10$ and $-10\leq x_2\leq 10$. The two-dimensional control input is bounded by $-5\leq u_i \leq 5,\ i = 1,2$. For simplicity, we assign constant weights to edges with certain existence. The constant $\gamma$ in \eqref{eq:unknown weight} is assigned to be 100. The sampled data points $N$ within each polytope in Algorithm 2 is set to be 100. To present the results, Fig. \ref{fig:1} illustrates details about the state-space partitioning, the predicted reachability-graranteed graph at the moment when the robot enters the target polytope, while Fig. \ref{fig:2} is the trajectory of the mobile robot operating in the environment. The environment is constructed in the MuJoCo simulator by defining a terrain geometry featuring uneven and bumpy surfaces. Surface properties, such as friction coefficients, are specified in the MuJoCo XML configuration to realistically simulate varying ground conditions and their effects on robot dynamics.

As shown in the center panel of Fig. \ref{fig:1}, the computed facet reachability involves not only grid cells with explored dynamics, but also predicts facet reachability in regions with unknown dynamics around the explored path. It also shows that our algorithm identifies a feasible path from the initial to the target.

\section{CONCLUSION AND FUTURE WORK} \label{sec:conclusion}
We propose a hybrid motion planning and control framework for robots operating in unknown dynamics that combines state space partitioning, affine system identification, predictive graph search and reach control synthesis. We showcase our strategy by demonstrating it on a mobile robot operating in an unknown environment.

Currently our task specification is confined to addressing the classical reach problem, which limits its capacity to capture complex robotic behaviors in unknown environments. In future work, we plan to extend our framework by integrating formal methods, such as linear temporal logic, to model a broader range of robotic interactions and behaviors. We also plan to rigorously derive bounds quantifying the discrepancy between the dynamics obtained through affine system identification and the true nonlinear dynamics as well as integrating them to the motion planning framework.




\section*{APPENDIX}
\subsection{Affine Dynamic Difference Bounds}\label{APP:Bounds}
Consider two affine systems \eqref{piecewise affine} obtained by linearizing the nonlinear system \eqref{dynamical system} at the operating points $(x_{1}, u_e = 0)$ and $(x_{2}, u_e = 0)$, yielding the affine representations $(A_1, B_1, c_1)$ and $(A_2, B_2, c_2)$ as in \eqref{piecewise affine}. Our objective is to rigorously bound the differences between these affine approximations. Namely, we seek to get $\varepsilon_A$, $\varepsilon_B$, $\varepsilon_c$, such that $\|A_2 - A_1\|\leq \varepsilon_A$, $\|B_2 - B_1\| \leq \varepsilon_B$, $\|c_2 - c_1\| \leq \varepsilon_c$.

From Assumption \ref{Lip ass}, one can obviously get
\begin{align}\small
    \|A_2 - A_1\| &= \|\nabla_xf(x_{2}) - \nabla_xf(x_{1})\| \leq L_{df}\|x_{2}-x_{ 1}\|,\\\small\label{eq:Ldf}
    \|B_2 - B_1\| &= \|g(x_{2}) - g(x_{1})\| \leq L_{g}\|x_{2} - x_{1}\|.
\end{align}
Thus, one can conclude that $\varepsilon_A = L_{df}\|x_{2}-x_{1}\|$, $\varepsilon_B = L_{g}\|x_{2} - x_{1}\|$.

To derive the bound for $\|c_2 - c_1\|$, we need to firstly derive the bound for $\|f(x_{2}) - f(x_{1})\|$ from a variation of \eqref{eq:Ldf}.
\begin{equation}\label{eq2}
\scalebox{0.87}{$\begin{aligned} 
    & f(x_{2}) - f(x_{1}) = \int_0^1\nabla_xf(x_{1}+t(x_{2}-x_{1}))(x_{2}-x_{1})dt\\
        \Rightarrow&\|f(x_{2}) - f(x_{1})\|\leq \int_0^1\|\nabla_xf(x_{1}+t(x_{2}-x_{1}))\|\|(x_{2}-x_{ 1})\|dt\\
        \Rightarrow&\|f(x_{2}) - f(x_{1})\|\leq \|(x_{2}-x_{ 1})\|\int_0^1\|\nabla_xf(x_{1}+t(x_{2}-x_{1}))\|dt.
    \end{aligned}$}
\end{equation}
For the term $\|\nabla_xf(x_{1}+t(x_{2}-x_{1}))\|$, we have
\begin{equation}\label{eq1}
\begin{aligned}
    \|\nabla_xf(x_{1}+t(x_{2}-x_{1}))\|\leq& \|\nabla_xf(x_{1}+t(x_{2}-x_{1}))\\
    &-\nabla_xf(x_1)\| + \|\nabla_x f(x_1)\|\\
    \leq& L_{df}t\|x_2 - x_1\| + \|A_1\|.
\end{aligned}
\end{equation}
Thus, plugging \eqref{eq1} into \eqref{eq2}, we have
\begin{equation}
    \begin{aligned}
        \|f(x_2) - f(x_1)\| \leq& \|x_2 - x_1\|\int_0^1(L_{df}t\|x_2 - x_1\|\\
        &+ \|x_2 - x_1\|)dt\\
        =& \|A_1\|\|x_2 - x_1\|+\frac{1}{2}L_{df}\|x_2-x_1\|^2.
    \end{aligned}
\end{equation}
Hence, for $\|c_2 - c_1\|$, we have
\begin{equation}
\scalebox{0.95}{$\begin{aligned}
           \|c_2 - c_1\| =& \|f(x_2) - f(x_1) + A_1 x_1 - A_2 x_2\|\\
     =& \|f(x_2) - f(x_1) + A_1 x_1 - A_1 x_2 + A_1 x_2- A_2 x_2\|\\
     =& \|f(x_2) - f(x_1) + A_1 (x_1 - x_2) + (A_1-A_2) x_2\|\\
     \leq& \|f(x_2) - f(x_1)\| + \|A_1\|\|x_2 - x_1\| \\
     &+ \|A_1 - A_2\|\|x_2\|\\
     \leq& 2\|A_1\|\|x_2 - x_1\| + \frac{1}{2}L_{df}\|x_2 - x_1\|^2 \\
     &+ L_{df}\|x_2 - x_1\|\|x_2\|.  
    \end{aligned}$}
\end{equation}
Hence, one can conclude that $\varepsilon_c = 2\|A_1\|\|x_2 - x_1\|+\frac{1}{2}L_{df}\|x_2 - x_1\|^2 + L_{df}\|x_2-x_1\|\|x_2\|$.

\subsection{Affine System Identification}
We implement an affine system identification procedure as Algorithm 2. The core approach involves continuously stimulating the system at very small time intervals to capture the corresponding system outputs (velocity), followed by applying linear regression to estimate the parameters of the affine system. With arbitrarily small control inputs and an arbitrarily small time step $T$, the system identification provides the dynamics with an arbitrarily small error \cite{ornik2019control}.

\begin{algorithm}[h]\label{alg:sysid}
\caption{System Identification Process}
\begin{algorithmic}[1]
\REQUIRE Initial state $x_{\text{init}} \in \mathbb{R}^n$, time step $T$, number of iterations $N$
\STATE \textbf{Initialize:}
\[
\begin{array}{l}
 \dot{x}\_list \gets \emptyset, \quad X\_list \gets \emptyset, \quad x \gets x_{\text{init}}
\end{array}
\]
\FOR{$i = 1$ \TO $N$}
    \STATE Generate small random control input $u=u_{rand}\in \mathbb{R}^m$
    \STATE Obtain the $\dot x$ and $x_{\text{new}}$ from the environment
    \STATE Append $\dot{x}$ to $\dot{x}\_list$
    \STATE Append the extended state vector $[x^\top, u^\top, 1]^\top$
    to $X\_list$
    \STATE Update the state: $x \gets x_{\text{new}}$
\ENDFOR
\STATE Estimate the system dynamics using least-squares:
\[
[A, B, c] \gets \dot x_{list} \, X_{list}^T \, \bigl( X_{list} \, X_{list}^T \bigr)^{-1}
\]
\RETURN $A$, $B$, $c$
\end{algorithmic}
\end{algorithm}

\subsection{Proof of Proposition 1}
For $i \in I_+$, the constraints are represented by $n_{l' i}^\top (\bar A_{l'}v_{l' j} + \bar B_{l'}u_{l'j} + \bar c_{l'})>0$. We seek to provide a lower bound of $n_{l' i}^\top (\bar A_{l'}v_{l' j} + \bar B_{l'}u_{l'j} + \bar c_{l'})$ with known $(\bar A_l, \bar B_l, \bar c_l)$ and dynamics difference bounds expressed by $\|\bar A_l - \bar A_{l'}\|\leq \varepsilon_A$, $\|\bar B_l - \bar B_{l'}\|\leq \varepsilon_B$, $\|\bar c_l - \bar c_{l'}\|\leq \varepsilon_c$. For term $n_{l'i}^\top (\bar A_{l'}v_{l'j} + \bar c_{l'})$,
\begin{equation}
    \scalebox{0.93}{$\begin{aligned} 
           n_{l'i}^\top& (\bar A_{l'}v_{l'j} + \bar c_{l'}) = n_{l'i}^\top [(\bar A_l+\bar A_{l'} - \bar A_l)v_{l'j} + (\bar c_{l}+\bar c_{l'}-\bar c_l)]\\
        &\leq n_{l'i}^\top (\bar A_l v_{l' j}+\bar c_l) + \|n_{l'i}\|(\|\bar A_{l'}-\bar A_l\|\|v_{l'j}\|+\|\bar c_{l'}-\bar c_l\|)\\
        &\leq n_{l'i}^\top (\bar A_l v_{l' j}+\bar c_l) + \|n_{l'i}\|(\varepsilon_A \|v_{l'j}\|+\varepsilon_c)
    \end{aligned}$}
\end{equation}
For term $n_{l'i}^\top \bar B_{l'} u_{l'j}$,
\begin{equation}
    n_{l'i}^\top \bar B_{l'} u_{l'j} = \begin{bmatrix}
        n_{l'i}^\top\bar B_{l'}^1\ \cdots\ n_{l'i}^\top\bar B_{l'}^m
    \end{bmatrix}\begin{bmatrix}
        u_{l'j}^1\\
        \cdots\\
        u_{l'j}^m
    \end{bmatrix},
\end{equation}
where $\bar B_{l'}^i,i=1\ldots m$, represents the $i$-th column of $\bar B_{l'}$. If $u_{l'j}^i>0$,
\begin{equation}
    \begin{aligned}
        n_{l'i}^\top\bar B_{l'}^iu_{l'j}^i& = n_{l'i}^\top(\bar B_{l}^i + \bar B_{l'}^i-\bar B_{l}^i) u_{l'j}^i\\
        &\ge (n_{l'i}^\top\bar B_{l}^i - \|n_{l'i}^\top\| \|\bar B_{l}^i-\bar B_{l'}^i\| )u_{l'j}^i\\
        &\ge (n_{l'i}^\top\bar B_{l}^i - \|n_{l'i}^\top\| \varepsilon_B )u_{l'j}^i.
    \end{aligned}
\end{equation}
Conversely, if $u_{l'j}^i\leq0$, $n_{l'i}^\top\bar B_{l'}^iu_{l'j}^i \ge (n_{l'i}^\top\bar B_{l}^i + \|n_{l'i}^\top\| \varepsilon_B )u_{l'j}^i$. Integrating the above, we obtain the following expression:
\begin{equation}
        \scalebox{0.92}{$\begin{aligned} 
 \bar B_{l'}u_{l'j} \ge  \left\{n^\top_{l'i}\bar B_{l}-\left[\frac{u_{l'j}^1}{|u_{l'j}^1|}\varepsilon_B\|n_{l'i}\|\ \cdots\ \frac{u_{l'j}^m}{|u_{l'j}^m|}\varepsilon_B\|n_{l'i}\|\right ] \right\}u_{l'j},
    \end{aligned}$}
\end{equation}
where we use $u_{l'j}^i/|u_{l'j}^i|$ to denote the sign of $u_{l'j}^i$. Hence, if 
\begin{equation}
    \bar B_{l'}^-u_{l'j}> -n_{l'i}^\top \bar A_{l} v_{l'j} - n_{l'i}^\top \bar c_{l} + \|n_{l'i}\|(\varepsilon_A\|v_{l'j}\| + \varepsilon_c)
\end{equation}
holds, we can conclude that
\begin{equation}
    \begin{aligned}
        \bar B_{l'}u_{l'j} &\ge B_{l'}^-u_{l'j}\\
        &> -n_{l'i}^\top \bar A_{l} v_{l'j} - n_{l'i}^\top \bar c_{l} + \|n_{l'i}\|(\varepsilon_A\|v_{l'j}\| + \varepsilon_c)\\
        &\ge n_{l' i}^\top (\bar A_{l'} v_{l' j} + \bar c_{l'}) 
    \end{aligned}
\end{equation}
holds. By rearranging the terms, the first inequality in \eqref{narrow ineq} holds. We analogously derive the case of $i\in I_-$ by simply enlarging the left-hand side and reducing the right-hand side of the inequality $n_{l'i}^\top \bar B_{l'}u_{l'j} \leq n_{l'i}^\top(\bar A_{l'} v_{l'j} + \bar c_{l'})$. The details are omitted for simplicity.

Thus, we can conclude that for any given $u_{l'j}$, if it lies within the feasible range specified in \eqref{narrow ineq}, then it will also lie within the feasible range defined by \eqref{eq: simple condition}.


\balance
\bibliographystyle{IEEEtran}
\bibliography{ref.bib}

\addtolength{\textheight}{-12cm}   

\end{document}